\documentclass[runningheads]{llncs}
\usepackage[T1]{fontenc}
\usepackage{graphicx,verbatim}
\usepackage[colorlinks=true]{hyperref}
\usepackage{enumitem}
\usepackage{amsmath}
\usepackage{stmaryrd}
\usepackage{bbold}
\usepackage{float}
\usepackage{makecell}
\usepackage{booktabs}

\usepackage{colortbl}
\definecolor{Gray}{gray}{0.9}

\begin{document}
\title{Distilling foundation models for robust and efficient models in digital pathology}
\titlerunning{H0-mini}
%

\author{Alexandre Filiot$^{1, \dagger, *}$, Nicolas Dop$^*$, Oussama Tchita$^1$, Auriane Riou$^1$, Rémy Dubois$^1$, Thomas Peeters$^2$, Daria Valter$^2$, Marin Scalbert$^2$, Charlie Saillard$^2$, Geneviève Robin$^1$, Antoine Olivier$^{1, \dagger}$.}  
\authorrunning{\authorrunning{A. Filiot et al.}}
\institute{$^1$ Owkin, Inc, $^2$ Bioptimus, Inc. \\
{$\dagger$ Corresponding authors.} \\
{$^*$ Equal contribution.} \\
    {\texttt{\{alexandre.filiot, oussama.tchita, auriane.riou, remy.dubois genevieve.robin,
antoine.olivier\}@owkin.com}, \texttt{\{thomas.peeters, dasha.walter, marin.scalbert, charlie.saillard\}@bioptimus.com, nicolas.dop@student-cs.fr}}
}

\maketitle              
\begin{abstract}

In recent years, the advent of foundation models (FM) for digital pathology has relied heavily on scaling the pre-training datasets and the model size, yielding large and powerful models. While it resulted in improving the performance on diverse downstream tasks, it also introduced increased computational cost and inference time. In this work, we explore the distillation of a large foundation model into a smaller one, reducing the number of parameters by several orders of magnitude. Leveraging distillation techniques, our distilled model, H0-mini, achieves comparable performance to large FMs at a significantly reduced inference cost on HEST and EVA public benchmarks. Additionally, we conduct robustness analyses on the PLISM dataset and a multi-scanner, multi-staining private breast cancer cohort. We demonstrate that our distilled model reaches excellent robustness to variations in staining and scanning conditions, significantly outperforming other state-of-the art models. This opens new perspectives to design lightweight and robust models for digital pathology, without compromising on performance. We publicly release H0-mini\footnote{\url{https://huggingface.co/bioptimus/H0-mini}} along with \texttt{plismbench}\footnote{Available at \url{https://github.com/owkin/plism-benchmark}.}, the first robustness benchmark of pathology foundation models based on the PLISM dataset.


\end{abstract}
\section{Introduction}

In recent years, representation learning has revolutionized computational pathology (CPath) by providing efficient models that can serve for a variety of tasks, including biomarker prediction, gene expression prediction, whole slide image (WSI) and tissue classification or survival analysis \cite{song2023}. Modern computational pathology frameworks typically rely on foundation models \cite{Chen_Ding_Lu_Williamson_Jaume_Chen_Zhang_Shao_Song_Shaban_et_al._2023,Filiot_Ghermi_Olivier_Jacob_Fidon_Kain_Saillard_Schiratti_2023,Filiot_Jacob_Mac_Kain_Saillard_2024,Ma_Guo_Zhou_Wang_Xu_Cai_Zhu_Jin_Lin_Jiang_et_al._2024,Nechaev_Pchelnikov_Ivanova_2024,Saillard_Jenatton_Llinares-López_Mariet_Cahané_Durand_Vert_2024,Xu_Usuyama_Bagga_Zhang_Rao_Naumann_Wong_Gero_González_Gu_et_al._2024,Zimmermann_Vorontsov_Viret_Casson_Zelechowski_Shaikovski_Tenenholtz_Hall_Klimstra_Yousfi_et_al._2024}, and share the common idea to leverage representation models (or feature extractors), able to map small patches of tissues (or tiles) to a lower dimensional space by computing features (or embeddings). Due to label scarcity at the tile-level, recent pathology feature extractors have been pretrained using self-supervised learning (SSL) methods such as DINO, iBOT or DINOv2 \cite{Caron_2021,Oquab_Darcet_Moutakanni_Vo_Szafraniec_Khalidov_Fernandez_Haziza_Massa_El-Nouby_et_al._2023,Zhou_Wei_Wang_Shen_Xie_Yuille_Kong_2022} and have become a cornerstone of modern CPath frameworks. Recent studies have mostly focused on scaling the pretraining dataset and the model size, resulting in large models called foundation models (FMs) whose size can reach more than one billion parameters, pretrained in up to a few million WSIs using the DINOv2 framework \cite{Saillard_Jenatton_Llinares-López_Mariet_Cahané_Durand_Vert_2024,Xu_Usuyama_Bagga_Zhang_Rao_Naumann_Wong_Gero_González_Gu_et_al._2024,Zimmermann_Vorontsov_Viret_Casson_Zelechowski_Shaikovski_Tenenholtz_Hall_Klimstra_Yousfi_et_al._2024}. This yielded spectacular improvements in performance on a variety of downstream tasks, yet introduced an increased computational cost.  Besides, the robustness of FMs to variations in sample preparation or digitization has not attracted the same attention as their performance, which is critical to the clinical deployment of CPath workflows \cite{US_Food_and_Drug_Administration_2020}. Some datasets and benchmarks have been proposed to investigate this aspect \cite{Lee_Lim_Byeon_Kwak_2024,Ochi_Komura_Onoyama_Shinbo_Endo_Odaka_Kakiuchi_Katoh_Ushiku_Ishikawa_2024,Wölflein_Ferber_Meneghetti_Nahhas_Truhn_Carrero_Harrison_Arandjelović_Kather_2023}.

To jointly tackle the issue of the additional computational cost while promoting the robustness of the models, this study proposes to investigate the distillation of large FMs. Starting from a recent foundation model, H-Optimus-0 \cite{Saillard_Jenatton_Llinares-López_Mariet_Cahané_Durand_Vert_2024}, a Vision Transformer (ViT)-giant \cite{Dosovitskiy_Beyer_Kolesnikov_Weissenborn_Zhai_Unterthiner_Dehghani_Minderer_Heigold_Gelly_et_al._2021} with more than one billion parameters, we investigate its distillation into a smaller ViT-Base of 86 million parameters. On several public benchmarks, the distilled model, H0-mini, demonstrates competitive performance with significantly larger state-of-the-art models. Additionally, leveraging the PLISM dataset \cite{Ochi_Komura_Onoyama_Shinbo_Endo_Odaka_Kakiuchi_Katoh_Ushiku_Ishikawa_2024} and a private multi-scanner, multi-staining dataset for breast cancer biomarkers prediction, we show that the distilled model significantly outperforms all other foundation models reported in the literature in its robustness to variations in staining and scanning conditions. This uncovers new perspectives to design robust CPath models in view of their adoption in clinical practice. 

\subsubsection{Related work.} Distillation is a well known machine learning technique that consists in supervising a student model by a teacher model \cite{Hinton_Vinyals_Dean_2015}. In the context of building image foundation models, it has been shown to be more efficient to distill a large model into a smaller one rather than training the small model from scratch \cite{Oquab_Darcet_Moutakanni_Vo_Szafraniec_Khalidov_Fernandez_Haziza_Massa_El-Nouby_et_al._2023}. In \cite{Duval_Misra_Ballas_2023}, a simple method is proposed to perform the distillation, and variations corresponding to the various SSL frameworks are investigated such as RoB-DINO and RoB-iBOT. While distillation techniques are well established in computer vision in general, their application to FMs for digital pathology remains relatively scarce. In GPFM \cite{Ma_Guo_Zhou_Wang_Xu_Cai_Zhu_Jin_Lin_Jiang_et_al._2024}, 3 foundation models (CONCH \cite{conch2024}, Phikon and UNI) are simultaneously distilled into a student model in addition to DINOv2 losses. In \cite{Zimmermann_Vorontsov_Viret_Casson_Zelechowski_Shaikovski_Tenenholtz_Hall_Klimstra_Yousfi_et_al._2024}, a ViT-Small model Virchow2G-Mini is introduced, resulting from the distillation of Virchow2G (a ViT-Giant) on one billion tiles.

\section{Material and methods}
\subsection{Pre-training setup}
\subsubsection{Distillation setup.} In this study, H-Optimus-0 (H0, \cite{Saillard_Jenatton_Llinares-López_Mariet_Cahané_Durand_Vert_2024}) is considered as the teacher model. We then follow the general methodology described in \cite{Oquab_Darcet_Moutakanni_Vo_Szafraniec_Khalidov_Fernandez_Haziza_Massa_El-Nouby_et_al._2023}. More precisely, for an image $x$, let $x_1$ and $x_2$ denote two augmented views of $x$. Additionally, for $i \in \llbracket 1, 2 \rrbracket$, we denote by $z_i^{(t)}$ (resp. $z_i^{(s)}$) the class tokens output by the teacher (resp. student) model for image $x_i$. For $i \in \llbracket 1, 2 \rrbracket$ and for a patch $p$, we also denote by $z_{i,p}^{(t)}$ (resp. $z_{i,p}^{(s)}$) the patch tokens output by the teacher (resp. student) for the image $x_i$. We pass the teacher and student class tokens (resp. patch tokens) through the corresponding DINO (resp. iBOT) head. $h_i$ denotes the vector of prototype scores output from the head projection of $z_i$. Taking inspiration from \cite{Duval_Misra_Ballas_2023}, distillation is performed by combining two objectives.

\paragraph{DINO objective.} In this setting, only the class scores are used to perform the distillation. Let $H$ denote the cross-entropy loss. The corresponding loss function $L_{\text{dino}}$ is defined as 
    \begin{equation}
        L_{\text{dino}} := \bigg(H(h_1^{(t)}, h_2^{(s)}) + H(h_2^{(t)}, h_1^{(s)})\bigg)/2.
    \end{equation}

\paragraph{IBOT objective.} The iBOT objective extends the distillation by incorporating patch scores supervision. This loss is defined as the following 
    \begin{equation}
        L_{\text{ibot}} := \frac{1}{2P} \sum_{p=1}^P \sum_{j=1}^2 H(h_{j, p}^{(t)}, h_{j, p}^{(s)}),
    \end{equation}
where $P$ denotes the total number of patches. Unlike \cite{Duval_Misra_Ballas_2023} we do not apply masking of the patches and the iBOT loss is applied for all patch scores.

Finally, for all experiments, we keep a spare exponential moving average (EMA) of the student as in \cite{Duval_Misra_Ballas_2023}, we remove the stochastic depth and Koleo regularization. To speed up the distillation, mixed precision is used as in \cite{Oquab_Darcet_Moutakanni_Vo_Szafraniec_Khalidov_Fernandez_Haziza_Massa_El-Nouby_et_al._2023}. H0-mini was trained for 105,000 iterations and a batch size of 2,048 using 128 Nvidia V100 32Go for a total of 4,350 gpu hours. For all downstream experiments, we used the spare EMA of the student as the feature extractor. The main hyperparameters used in the distillation are given in appendix \ref{supp:hyperparameters}.

\subsubsection{Pre-training datasets.} We perform the distillation on a dataset of 43M tiles (224x224 at 20x magnification) extracted from 6,093 TCGA slides covering 16 cancer sites. This allows for a fair comparison with Phikon, a ViT-B feature extractor pretrained from scratch on the same dataset with iBOT. Compared to the typical sizes of pre-training datasets, it stands out by its relatively small size. In particular, we note the difference with the pretraining setup proposed in \cite{Zimmermann_Vorontsov_Viret_Casson_Zelechowski_Shaikovski_Tenenholtz_Hall_Klimstra_Yousfi_et_al._2024}, where the distillation is performed on a 1B-tile dataset.

\subsection{Evaluation setup}
\subsubsection{Performance benchmarks.} 
To rigorously evaluate model performance, we use two publicly available benchmarks\footnote{The HEST benchmark is available at \url{https://github.com/mahmoodlab/HEST}, and the EVA benchmark is available at \url{https://kaiko-ai.github.io/eva/main/}.}. The EVA benchmark \cite{Gatopoulos_Känzig_Moser_Otalora_2024} consists in 4 patch-level classification tasks, 2 patch-level segmentation tasks, and 2 slide-level classification tasks. For each category of tasks, EVA’s evaluation protocol is fixed to allow for a fair evaluation across feature extractors (\textit{e.g.}, learning rate, batch size or schedulers). We also leverage the HEST-Benchmark \cite{Jaume_Doucet_Song_Lu_Almagro-Pérez_Wagner_Vaidya_Chen_Williamson_Kim_et_al._2024} which collects gene expression prediction tasks for 9 different indications. For each task, a subset of 50 highly variable genes is considered. To account for various embedding dimensions between feature extractors, we follow the recommended evaluation procedure from HEST-Benchmark, fitting a ridge regression on top of a PCA reduction with 256 components. Additionally, and following the recommended methodology in \cite{Zimmermann_Vorontsov_Viret_Casson_Zelechowski_Shaikovski_Tenenholtz_Hall_Klimstra_Yousfi_et_al._2024}, we use the concatenation of the class token and the mean over all patch tokens as embeddings of our distilled model.

\subsubsection{PLISM robustness dataset.} To evaluate model robustness, we first use the public PLISM \cite{Ochi_Komura_Onoyama_Shinbo_Endo_Odaka_Kakiuchi_Katoh_Ushiku_Ishikawa_2024} dataset which consists of 46 human tissue types stained using 13 different H\&E conditions, and captured using 7 WSI scanners (Figure \ref{fig:plism_main_result}). This results in 91 WSIs later registered using the Elastix software \cite{Klein2009-kv} and tessellated into 16,278 tissue tiles each. We make this processed dataset publicly available\footnote{The processed PLISM dataset is made available at \url{https://huggingface.co/datasets/owkin/plism-dataset}}.

Cosine similarity and top-$k$ accuracy are used to evaluate the robustness of the feature extractors across scanner and staining combinations. Cosine similarity is assessed on features extracted from tissue tiles averaged over all matching tile pairs. We note that this metric was proposed in \cite{Wölflein_Ferber_Meneghetti_Nahhas_Truhn_Carrero_Harrison_Arandjelović_Kather_2023} to quantify the robustness of feature extractors to stain normalization. Top-k accuracy computes the percentage of tiles from one slide whose matching tile on the other slide ranks among the $k$ closest tiles (by cosine similarity) when compared to all other tiles from both slides. Finally, metrics are aggregated across slide pairs. Specific robustness to scanner (resp. staining) is assessed on all fixed-staining (resp. scanner) cross-scanner (resp. staining) pairs. We also report overall robustness on cross-stainings, cross-scanners slide pairs. Additional details on the implementation of those metrics are given in appendix \ref{supp:plism_metrics_details}. 

\subsubsection{BreastBiomarker (BreastBm) performance \& robustness benchmark.} 
\begin{figure}[h!]
\includegraphics[width=\textwidth]{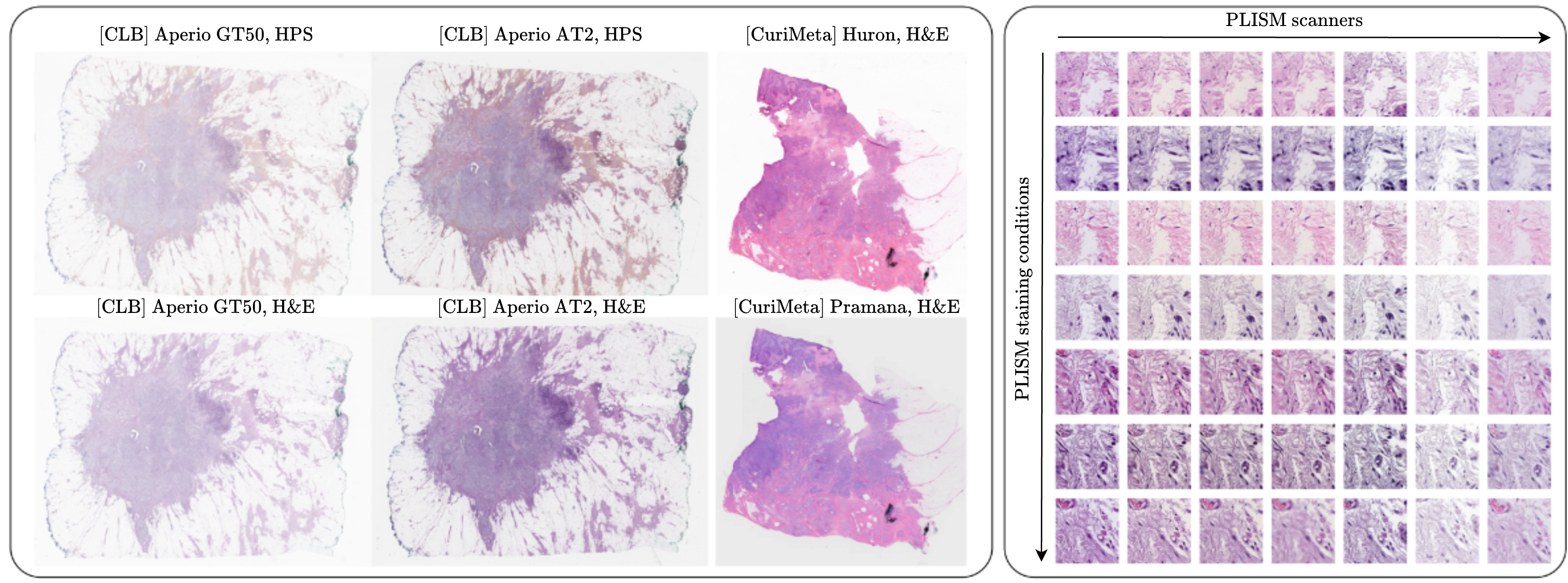}
\caption{Visualizations of BreastBm (left) and PLISM (right) datasets. For PLISM, we only display 7 of the 13 different stainings on the y-axis.} \label{fig:plism_main_result}
\end{figure}

We also evaluate the downstream robustness of H0-Mini to variations in scanner and staining in the context of breast cancer biomarker prediction. Slide-level tasks aim at predicting the positivity of estrogen receptor status (ER), progesterone receptor status (PR), human epidermal growth factor receptor 2 status (HER2) and germline BRCA1/BRCA2 mutations (gBRCA). We leverage 3 private cohorts for robustness evaluation (see Figure \ref{fig:plism_main_result}). The “GR” (resp. “CLB”) cohorts consists of 671 (resp. 353) tumor blocks providing 2 consecutive slides: one stained with Hematoxylin Eosin Saffron (HES) (resp. archival Hematoxylin Phloxine Saffron, HPS) and one with H\&E. All GR (resp. CLB) slides are scanned using an Olympus VS200 (resp. Aperio GT450 and Aperio AT2). GR H\&E slides are also scanned with Aperio ScanScope and Hamamatsu S60. This yields 6 possible subcohort pairs for GR and for CLB. The CuriMeta cohort consists of 208 H\&E slides scanned with Pramana and Huron scanners (1 pair).

For all endpoints and to best isolate the impact of feature extractors, all downstream models use a mean pooling architecture: a logistic regression is trained on the average of tile-level feature representations. Performance is assessed using AUC while robustness is evaluated via the Concordance Correlation Coefficient (CCC) between WSI-level predictions from different stainings and scanners for the same tumor blocks. For ER and PR (resp. HER2) endpoints, training was performed on 1,109 WSIs from TCGA-BRCA and evaluation on GR, CLB and CuriMeta cohorts (resp. GR cohort). Due to labels unavailability in TCGA, predictive models for the gBRCA endpoint were trained on GR and validated on CLB and CuriMeta.

\section{Results}
\subsection{Performance evaluation}

In Table \ref{tab:eva_results}, we report the results on the EVA benchmark. As suggested by its authors, we report the average results without the BACH task, as the spatial resolution of the images differs from the other tasks and therefore tend to favor mixed-magnification models such as Virchow2 or the Kaiko models \cite{kaikoFM2024}. Even though the results tend to saturate (6 models show an average performance between 0.78 and 0.79), H0-mini is competitive with the other state-of-the-art models while being much smaller. A noticeable exception is the Kaiko-B/8 model reaching equivalent performance (but with a smaller patch size which results in more computations). This shows that the distilled model can perform well on a variety of tasks, including patch classification, WSI classification or segmentation.

\begin{table}[h!]
\centering
\caption{EVA results. For all models, input embeddings are only the [CLS] tokens. Balanced accuracy (resp. “MonaiDiceScore”) is reported for classification (resp. segmentation) tasks. ($^*$) results may be over-estimated due to an overlap between downstream and pre-training datasets.}
\label{tab:eva_results}
{\fontsize{8}{\baselineskip}\selectfont
\begin{tabular}{l c c c c c c c c c c c}
\toprule
 \textbf{} & & \multicolumn{4}{c}{\textbf{Patch-level classification}} & \multicolumn{2}{c}{\makecell{\textbf{Slide-level} \\ \textbf{classification}}} & \multicolumn{2}{c}{\textbf{Segmentation}} & \multicolumn{1}{c}{\textbf{Mean}} \\
\textbf{Model} & \textbf{Size} & \textbf{Bach} & \textbf{Crc} & \textbf{Mhist} & \textbf{Pcam} & \textbf{Cam16} & \makecell{\textbf{Panda}} & \textbf{Consep} & \textbf{Monusac} & \makecell{\textbf{All}} \\ 
Virchow2 & 632M & \underline{0.880} & \textbf{0.966} & \textbf{0.858} & 0.936 & \textbf{0.864} & 0.642 & 0.630 & 0.663 & \textbf{0.794} \\
UNI2-h  & 682M & \textbf{0.914} & \underline{0.965} & 0.820 & \underline{0.949} & \underline{0.855} & \textbf{0.672} & 0.632 & 0.642 & \underline{0.791} \\
H0 & 1,100M  & 0.758 & 0.958 & 0.839 & 0.942 & 0.820 & 0.645 & 0.637 & \textbf{0.679} & 0.789 \\
Gigapath & 1,100M  & 0.761 & 0.952 & 0.829 & 0.945 & 0.814 & 0.664 & 0.621 & 0.672 & 0.785 \\
GPFM & 307M  & 0.830* & 0.952 & 0.811 & 0.945 & 0.851* & 0.647* & 0.639 & 0.640 & 0.784 \\
UNI& 307M  & 0.797 & 0.947 & \underline{0.844} & 0.936 & 0.834 & 0.656 & 0.628 & 0.638 & 0.783 \\
\rowcolor{Gray} \textbf{H0-mini}& 86M   & 0.774 & 0.961 & 0.790 & 0.942 & 0.842 & \underline{0.667} & 0.629 & 0.643 & 0.782 \\
Hibou L$_{16}$ & 307M  & 0.816 & 0.931 & 0.826 & \textbf{0.951} & 0.832 & 0.633 & \underline{0.642} & 0.658 & 0.782 \\
Kaiko B$_8$ & 86M  & 0.858 & 0.957 & 0.823 & 0.918 & 0.818 & 0.638 & \textbf{0.645} & \underline{0.675} & 0.782 \\
Phikon & 86M  & 0.722 & 0.936 & 0.799 & 0.922 & 0.797 & 0.640 & 0.629 & 0.644 & 0.767 \\
PhikonV2 & 307M  & 0.727 & 0.939 & 0.775 & 0.893 & 0.808 & 0.635 & 0.630 & 0.639 & 0.760 \\
\bottomrule
\end{tabular}
}
\end{table}

In Table \ref{tab:hest_results}, we report the results on the HEST benchmark. H0-mini also performs well on this benchmark highlighting the efficiency of knowledge distillation. The model significantly outperforms Phikon, its ViT-B equivalent trained from scratch with iBOT on the same pre-training data. Moreover, it surpasses on average much bigger foundation models such as Virchow2 or Gigapath.

\begin{table}[h!]
\centering
\caption{HEST results (Pearson correlations). For all models, [CLS] tokens are concatenated to the mean over all patch tokens to form the input embeddings.}
\label{tab:hest_results}
{\fontsize{8}{\baselineskip}\selectfont
\begin{tabular}{ l c c c c c c c c c c }
\toprule
\textbf{Model} & \textbf{Idc} & \textbf{Prad} & \textbf{Paad} & \textbf{Skcm} & \textbf{Coad} & \textbf{Read} & \textbf{Ccrcc} & \textbf{Luad} & \textbf{L. Idc} & \textbf{Mean} \\  

 UNI2-h  & \underline{0.6054} & 0.3753 & \textbf{0.5231} & \textbf{0.6829} & \textbf{0.3319} & \underline{0.2265} & 0.2662 & 0.5743 & \textbf{0.2743} & \textbf{0.4292} \\
H0  & \textbf{0.6106} & 0.3621 & \underline{0.5106} & \underline{0.6614} & 0.3089 & \textbf{0.2401} & \underline{0.2669} & \underline{0.5754} & 0.2664 & \underline{0.4224} \\ 
\rowcolor{Gray} \textbf{H0-mini}  & 0.5909 & 0.3633 & 0.5068 & 0.6125 & 0.2700 & 0.2047 & 0.2643 & 0.5633 & 0.2640 & 0.4044 \\ 
Virchow2 & 0.5971 & 0.3528 & 0.4778 & 0.6404 & 0.2580 & 0.2073 & 0.2604 & 0.5685 & 0.2568 & 0.4019 \\ 
Hibou L$_{16}$ & 0.5945 & 0.3231 & 0.4758 & 0.6059 & \underline{0.3128} & 0.1823 & \textbf{0.2777} & 0.5720 & 0.2490 & 0.3992 \\
Gigapath & 0.5707 & \textbf{0.3841} & 0.4920 & 0.5823 & 0.3076 & 0.186 & 0.2277 & 0.5579 & 0.2499 & 0.3952 \\ 
GPFM & 0.5796 & 0.3733 & 0.4686 & 0.5839 & 0.2801 & 0.1769 & 0.2510 & 0.5522 & 0.2391 & 0.3940 \\
Kaiko B$_8$ & 0.5710 & \underline{0.3827} & 0.4727 & 0.5904 & 0.3105 & 0.1726 & 0.2664 & \textbf{0.5883} & 0.2362 & 0.3912 \\ 
UNI & 0.5851 & 0.3274 & 0.4882 & 0.6235 & 0.2583 & 0.1757 & 0.2463 & 0.5558 & 0.2576 & 0.3907 \\ 
PhikonV2 & 0.5677 & 0.3793 & 0.4771 & 0.5845 & 0.2561 & 0.1865 & 0.2607 & 0.5502 & 0.2476 & 0.3897 \\ 
Phikon & 0.5481 & 0.3452 & 0.4639 & 0.5555 & 0.2668 & 0.1667 & 0.2496 & 0.5679 & 0.2387 & 0.3780 \\
\bottomrule
\end{tabular}
}
\end{table}

Overall ranking and comparison between models is performed through a one-sided Wilcoxon sign rank test with Holm correction (at $\alpha=0.05$) on HEST and EVA combined (17 tasks). This places H0-mini at 4th position, outperformed by UNI2-h ($p < 10^{-4}$), H-Optimus-0 ($p=0.039$) and Virchow2 (n.s., $p=0.29$). H0-mini statistically outperforms UNI ($p=0.044$) and the Phikon models ($p < 10^{-3}$).

\subsection{Robustness evaluation} 

Table \ref{tab:plism_results} presents the robustness metrics evaluated on the PLISM dataset. H0-mini achieves high scores, outperforming all other state-of-the-art models. We also note that multi-modal extractors, such as CONCH, demonstrate high robustness. While there is a correlation between the cosine similarity metric and the retrieval metric, cosine similarity alone is not enough to assess the robustness of a foundation model. Models such as PLIP \cite{plip2023} and Phikon present high values of cosine similarity, while failing at the retrieval task. It should also be noted that robustness to staining is harder to achieve compared to scanner as staining variations may have an impact on biological morphology while scanning variations mostly impact color tints. Looking at top-10 accuracy, H0-Mini presents the highest robustness to scanning variations, as well as joint scanning and staining variations.

\begin{table}[h!]
\centering
\caption{PLISM results (cosine similarity and top-10 accuracy). For each metric, we report median and inter-quartile range over the corresponding slide pairs.}
\label{tab:plism_results}

{\fontsize{8}{\baselineskip}\selectfont

\begin{tabular}{lcccccc}
\toprule
& \multicolumn{2}{l}{\makecell{\textbf{Fixed-staining,} \\ \textbf{cross-scanner}}} & \multicolumn{2}{l}{\makecell{\textbf{Fixed-scanner,} \\ \textbf{cross-staining}}} & \multicolumn{2}{l}{\makecell{\textbf{Cross-staining,} \\ \textbf{cross-scanner}}} \\
\textbf{Model} & \textbf{Cosine} & \textbf{top-10} & \textbf{Cosine} & \textbf{top-10} & \textbf{Cosine} & \textbf{top-10 $\downarrow$} \\
\rowcolor{Gray} \textbf{H0-mini}  & 0.92 (0.06) & \textbf{0.86 (0.28)} & 0.83 (0.09) & \underline{0.32 (0.33)} & 0.79 (0.11) & \textbf{0.18 (0.25)} \\
H0  & 0.85 (0.08) & 0.74 (0.32) & 0.73 (0.11) & \textbf{0.33 (0.29)} & 0.67 (0.12) & \underline{0.17 (0.20)} \\
Virchow2 & 0.88 (0.06) & 0.61 (0.31) & 0.81 (0.09) & 0.31 (0.25) & 0.77 (0.10) & 0.16 (0.18) \\
CONCH  & \textbf{0.93 (0.04)} & \underline{0.75 (0.29)} & \underline{0.86 (0.05)} & 0.24 (0.21) & \underline{0.84 (0.06)} & 0.16 (0.16) \\
GigaPath  & 0.79 (0.11) & 0.59 (0.40) & 0.61 (0.13) & 0.12 (0.18) & 0.56 (0.14) & 0.05 (0.08) \\
UNI & 0.76 (0.09) & 0.53 (0.30) & 0.62 (0.13) & 0.17 (0.22) & 0.53 (0.13) & 0.05 (0.08) \\
UNI2-h & 0.76 (0.17) & 0.50 (0.56) & 0.68 (0.13) & 0.19 (0.25) & 0.56 (0.17) & 0.05 (0.10) \\
Kaiko B$_8$ & 0.87 (0.07) & 0.35 (0.46) & 0.81 (0.09) & 0.15 (0.20) & 0.75 (0.11) & 0.05 (0.09) \\
GPFM  & 0.80 (0.08) & 0.36 (0.34) & 0.69 (0.17) & 0.09 (0.18) & 0.58 (0.14) & 0.02 (0.04) \\
Hibou L$_{16}$ & 0.65 (0.10) & 0.06 (0.09) & 0.61 (0.16) & 0.03 (0.06) & 0.47 (0.14) & 0.008 (0.01) \\
PLIP  & \underline{0.92 (0.04)} & 0.05 (0.20) & \textbf{0.91 (0.05)} & 0.04 (0.10) & \textbf{0.87 (0.06)} & 0.004 (0.01) \\
Phikon  & 0.81 (0.08) & 0.13 (0.24) & 0.69 (0.19) & 0.02 (0.07) & 0.60 (0.18) & 0.004 (0.01) \\
PhikonV2  & 0.72 (0.09) & 0.06 (0.10) & 0.66 (0.18) & 0.03 (0.07) & 0.54 (0.14) & 0.003 (0.008) \\
\bottomrule
\end{tabular}
}
\end{table}

Finally, table \ref{tab:brca_results} presents the downstream results on the BreastBm dataset. H0-mini consistently and significantly outperforms H-Optimus-0 on each of the performance and robustness metrics ($p < 10^{-4}$) as assessed by multiple boot-strap-based comparison tests combined with harmonic mean p-value across subcohorts (resp. subcohort pairs) for AUC (resp. CCC). Considering that the features aggregation module is a simple average, BreastBm results suggest that robustness in the feature space translates to downstream robustness. 

\begin{table}[ht]
\centering
\caption{Performance (measured by AUC) and robustness (measured by CCC) results on the BreastBM dataset.}
\label{tab:brca_results}
{\fontsize{8}{\baselineskip}\selectfont

\begin{tabular}{llcccccc}
\toprule
& & \multicolumn{3}{l}{\makecell{\textbf{Performance metric: AUC}}}  & \multicolumn{3}{l}{\makecell{\textbf{Robustness metric: CCC}}}  \\

\textbf{Task} & \textbf{Model} & \textbf{Clb} & \textbf{Gr} & \textbf{Curimeta} & \textbf{Clb} & \textbf{Gr} & \textbf{Curimeta} \\

 gBrca & H0-mini & \textbf{0.69} (0.02) & - & \textbf{0.69} (0.01) & \textbf{0.92} (0.03) & - & \textbf{0.88} (0.84-0.91) \\
 & H0 & 0.66 (0.02) & - & 0.64 (0.05) & 0.42 (0.25) & - & 0.71 (0.65-0.78) \\

\rowcolor{Gray} ER & H0-mini & \textbf{0.95} (0.00) & \textbf{0.76} (0.02) & \textbf{0.88} (0.00) & 0.97 (0.02) & \textbf{0.82} (0.08) & \textbf{0.98} (0.97-0.98) \\
 \rowcolor{Gray}  & H0 & 0.95 (0.01) & 0.75 (0.03) & \textbf{0.88} (0.00) & \textbf{0.97} (0.01) & 0.72 (0.12) & 0.90 (0.87-0.95) \\

PR & H0-mini & \textbf{0.89} (0.01) & 0.87 (0.00) & \textbf{0.69} (0.01) & \textbf{0.96} (0.03) & \textbf{0.78} (0.10) & \textbf{0.98} (0.97-0.98) \\
 & H0 & 0.88 (0.01) & \textbf{0.88} (0.02) & \textbf{0.69} (0.01) & 0.92 (0.03) & 0.63 (0.18) & 0.94 (0.92-0.96) \\

\rowcolor{Gray} HER2 & H0-mini & - & \textbf{0.66} (0.02) & - & - & \textbf{0.74} (0.14) & - \\
\rowcolor{Gray}  & H0 & - & 0.65 (0.02) & - & - & 0.64 (0.18) & - \\
\bottomrule
\end{tabular}
}
\end{table}

\section{Discussion}

While the results presented in this work are promising, several limitations and perspectives can be noted. First, we note that the final projection head of the teacher model is required to perform the distillation. Whether the distillation can be performed without the projection head (for instance, by learning them during the pre-training as in \cite{Ma_Guo_Zhou_Wang_Xu_Cai_Zhu_Jin_Lin_Jiang_et_al._2024}) with equivalent downstream performance and robustness should be explored. Second, even though distilled models will be much cheaper to use for downstream tasks (both in time, cost or carbon footprint), they remain computationally demanding to train compared to training a similar model from scratch (1.7x longer for a ViT-Base) as it requires the teacher model. However, the resulting performance and robustness is greatly improved.

Finally, we mention some open questions. The successful distillation of large models into smaller ones raises a question on the intrinsic dimension of foundation models in digital pathology. Besides, we did not perform an extensive hyperparameter optimization. Whether some could be tailored for distillation remains to be explored (\emph{e.g.}, adding more global crops, reducing the student patch size, etc.). Additionally, we proposed several metrics to assess the robustness of foundation models, yet future work should be carried out to have a more comprehensive understanding of a model’s behavior \cite{dejong2025currentpathologyfoundationmodels}. Note that this may require high-quality multi-scanned, multi-stained datasets. We hope that releasing our processed version of PLISM will encourage further research in this direction.

\section{Conclusion}

This study shows that distilling a large foundation model results in a competitive smaller model. We show that this model can efficiently be leveraged for downstream applications closing the performance gap with larger foundation models. By releasing the model publicly, we hope our work will democratize the use of pathology foundation models for researchers with limited computational resources.
 
Besides, we show that distilled models present an additional robustness property, with their features having a better invariance to changes in scanners or staining conditions. Self-supervised learning is known to be an efficient way to address domain generalization in CPath \cite{Zamanitajeddin_Jahanifar_Xu_Siraj_Rajpoot_2024} and, to the best of our knowledge, this study is the first to highlight that distilling a large foundation model further improves the robustness of the resulting model. This enhanced robustness property lays the groundwork for future research, supporting the broader adoption of CPath models in clinical practice.

    

\begin{credits}
\subsubsection{\ackname} 
\paragraph{Computing resources.}
This work was granted access to the High-Performance Computing (HPC) resources of IDRIS under the allocations 2023-A0141012519, 2024-A0161012519 and 2024-GC011015442 made by GENCI.

\paragraph{Data access.}
The results presented here are in part based upon data generated by the TCGA Research Network: \url{https://www.cancer.gov/tcga}. Results on the gBRCA dataset are based on data generated by Gustave Roussy (GR) and Centre Léon Bérard (CLB) through PortrAIt (Owkin, Tribun Health, Cypath, GR, CLB, Unicancer), a French consortium advancing precision medicine. PortrAIt was funded by the French government as part of the France 2030 program, and the European Union – NextGenerationEU through the France Relance program An additional external data set from the United States was accessed through CuriMeta.

\end{credits}



%
%
%
\bibliographystyle{splncs04}
\bibliography{references}
%





\appendix
\section{Supplemental}

\subsection{Distillation hyperparameters}
\label{supp:hyperparameters}

We provide in Table \ref{tab:hyperparameters} the main hyperparameters used in the distillation.

\begin{table}[h!]
\centering
\caption{Main hyperparameters used in the distillation.}
\label{tab:hyperparameters}
\resizebox{\textwidth}{!}{
\begin{tabular}{l l l l}
\toprule
\\
\textbf{Optimization} & Warmup epochs & & 16 \\
 \textbf{Parameters} & Teacher temperature warmup epochs && 30 \\
 & Weight decay end value && 0.4 \\
 & Total batch size && 2,048 \\
 & Number of iterations && 105,000 \\ \\
\textbf{Model} & Patch size && 14 \\
 \textbf{Parameters} & Register tokens && 4 \\
 & Embedding dimension && 768 \\
 & Layers && 12 \\
 & Heads && 12 \\
 & MLP ratio && 4 \\
 & MLP activation && SwiGLU \\ \\
\textbf{Projection} & Heads prototypes && 131,072 \\
 \textbf{Heads} & DINO head bottleneck dim && 384 \\
 & iBOT head bottleneck dim && 256 \\ \\
\textbf{Hardware} & GPUs && 128 V100 32 Go \\
\bottomrule
\end{tabular}
}
\end{table}

\subsection{Mathematical details of robustness evaluation metrics}
\label{supp:plism_metrics_details}
\subsubsection{Cosine Similarity between Tiles}
The cosine similarity between two tiles is a measure of the similarity between their feature vectors. Let \( t_1 \) and \( t_2 \) represent the feature vectors of two tiles. The cosine similarity between \( t_1 \) and \( t_2 \) is defined as:
\begin{equation*}
    \text{Cosine Similarity}(t_1, t_2) = \frac{\langle t_1, t_2 \rangle}{\| t_1 \|_2 \| t_2 \|_2},
\end{equation*}
where
\begin{itemize}[label=$\bullet$] 
    \item \( \langle \cdot, \cdot \rangle \) denotes the canonical dot product,
    \item and \( \| \cdot \|_2 \) denotes the corresponding euclidean norm.
\end{itemize}

\subsubsection{Cosine Similarity for a Slide Pair}

For each slide pair \( (S_1, S_2) \), the Mean Cosine Similarity is computed by averaging the cosine similarities between corresponding tile pairs. Each tile pair \( (t_{i,1}, t_{i,2}) \) corresponds to a matched location, where \( t_{i,1} \) and \( t_{i,2} \) are the feature vectors extracted from slides \( S_1 \) and \( S_2 \), respectively. Let \( N \) denote the total number of tile pairs. The Mean Cosine Similarity between \( S_1 \) and \( S_2 \) is then:

\begin{equation*}
   \text{Mean Cosine Similarity}_{S_1, S_2} = \frac{1}{N} \sum_{i=1}^{N} \text{Cosine Similarity}(t_{i,1}, t_{i,2}).
\end{equation*}

\subsubsection{Top-k Accuracy for a Slide Pair}

For each tile \( t_{i, 1} \) from slide \( S_1 \), the cosine similarities between \( t_{i,1} \) and all other tiles from both \( S_1 \) and \( S_2 \) are computed. These tiles are then ranked based on their similarity. The Top-k Accuracy is defined as the fraction of tiles from \( S_1 \) whose corresponding tile from \( S_2 \) ranks among the top \( k \) closest tiles. Formally, the Top-k Accuracy from \( S_1 \) to \( S_2 \) is:

\begin{equation*}
   \text{Top-k Accuracy}_{S_1 \to S_2} = \frac{1}{N} \sum_{i=1}^{N} \mathbb{1} \left( \text{rank}_{S_1 \cup S_2}(t_{i,2}) \leq k \right),
\end{equation*}

where the rank of \( t_{i,2} \) in \(S_1 \cup S_2 \) is defined as: 
\[
\resizebox{\textwidth}{!}{$
\text{rank}_{S_1 \cup S_2}(t_{i,2}) = \left| \{ t \in (S_1 \cup S_2), t \neq t_{i,1}: \text{Cosine Similarity}(t_{i,1}, t) \geq \text{Cosine Similarity}(t_{i,1}, t_{i,2}) \} \right|.
$}
\]

\noindent The final Top-k Accuracy for the slide pair is obtained by averaging the Top-k Accuracy in both directions:
\begin{equation*}
  \text{Top-k Accuracy}_{S_1, S_2} = \frac{1}{2} \left( \text{Top-k Accuracy}_{S_1 \to S_2} + \text{Top-k Accuracy}_{S_2 \to S_1} \right).  
\end{equation*}

\end{document}